\documentclass[10pt, a4paper]{article}
\usepackage{lrec-coling2024} 

\usepackage{natbib}
\usepackage{multibib}
\makeatletter
\def\@mb@citenamelist{cite,citep,citet,citealp,citealt,citepalias,citetalias}
\makeatother
\newcites{languageresource}{~}

\usepackage{graphicx}
\usepackage{tabularx}
\usepackage{soul}
\usepackage{makecell}
\usepackage{adjustbox}

\newcolumntype{Y}{>{\centering\arraybackslash}X}

\usepackage{xcolor} \usepackage{hyperref}
 \definecolor{darkblue}{rgb}{0, 0, 0.5}
  \hypersetup{colorlinks=true, citecolor=darkblue, linkcolor=darkblue, urlcolor=darkblue}

\usepackage{xstring}
\usepackage{color}

\usepackage[utf8]{inputenc} % allow utf-8 input
\usepackage[T1]{fontenc}    % use 8-bit T1 fonts
\usepackage[arabic,english]{babel}

\title{Advancing the Arabic WordNet: Elevating Content Quality\\ \vspace*{.5\baselineskip}}

\name{Abed Alhakim Freihat\,$^{1}$, Hadi Khalilia\,$^{1,2,*}$, Gábor Bella\,$^{3}$, Fausto Giunchiglia\,$^{1}$}

\address{$^{1}$Department of Information Engineering and Computer Science, University of Trento, Italy \\
        $^{2}$Palestine Technical University – Kadoorie, Palestine\\
        $^{3}$Lab-STICC CNRS UMR 628, IMT Atlantique, Brest, France\\
        $^{1}$ \{abed.freihat, hadi.khalilia, fausto.giunchiglia\}@unitn.it,\\
        $^{2}$h.khalilia@ptuk.edu.ps\\
        $^{3}$gabor.bella@imt-atlantique.fr\\
}

\abstract{
High-quality WordNets are crucial for achieving high-quality results in NLP applications that rely on such resources. However, the wordnets of most languages suffer from serious issues of correctness and completeness with respect to the words and word meanings they define, such as incorrect lemmas, missing glosses and example sentences, or an inadequate, Western-centric representation of the morphology and the semantics of the language. Previous efforts have largely focused on increasing lexical coverage  while ignoring other qualitative aspects. In this paper, we focus on the Arabic language and introduce a major revision of the Arabic WordNet that addresses multiple dimensions of lexico-semantic resource quality. As a result, we updated more than 58\% of the synsets of the existing Arabic WordNet by adding missing information and correcting errors. In order to address issues of language diversity and untranslatability, we also extended the wordnet structure by new elements: \emph{phrasets} and \emph{lexical gaps}.
 \\ \newline \Keywords{Arabic, wordnet, quality, completeness, correctness, phraset, lexical semantics} }

\begin{document}

\maketitleabstract

\section{Introduction}
WordNets \cite{beckwith2021} are lexical databases that represent lemmas (lexemes, words) of a language, together with their meanings organised into a lexico-semantic network. Wordnets define meanings as sets of synonymous words called \emph{synsets}. Synsets are described by a gloss (e.g., a definition in a natural language that represents the synset meaning) as well as example sentences that clarify the usage of words in context. WordNets are used in many NLP applications, such as machine translation \cite{poibeau2017}, information retrieval \cite{nie2022}, or word sense disambiguation \cite{navigli2009}.

The English Princeton WordNet (PWN) \citeplanguageresource{miller1995}, as the first wordnet, has been adapted and employed as a foundation for constructing wordnets in other languages.

In general, WordNets are constructed using either the \emph{merge} or the \emph{expand model} \cite{vossen1998}. In the merge model, synsets are initially created from  pre-existing resources (e.g., dictionaries) in a language. Then, for translability into other languages, the synsets have to be aligned with equivalent English synsets in PWN. For example, the IndoWordNet \citeplanguageresource{bhattacharyya2010} was built following this model. In the expand model, PWN synsets are `localized' or `translated' into target languages. For example, the Polish WordNet \citeplanguageresource{piasecki2009} was constructed using this model. In either case, when mapping across languages, the PWN synsets (and thus the English language) are usually used as a pivot when translating words across languages.

Wordnets often suffer from quality issues, in a large part due to the use of automated and semi-automated methods for building them \cite{khalilia2021a,khalilia2021b}. In addition, mistakes can be hard to detect as most wordnets do not contain glosses or example sentences. The above are true of the existing Arabic wordnets. The first Arabic wordnet (AWN V1) was built following the expand model \citeplanguageresource{elkateb2006} and includes 9,618~synsets translated from PWN to modern standard Arabic. Its second version (AWN~V2) \citeplanguageresource{regragui2016} extended AWN~V1 to 11,269~synsets and was developed using a semi-automatic method and the expand model. As we show in our paper, both wordnets suffer from correctness and completeness issues, and lack glosses and examples. By correctness we refer to the accuracy of lemmas in representing the meaning of a synset, while completeness refers to the extent to which a synset includes all words that are synonymous based on the synset meaning. For example, without an Arabic gloss and example sentences, it is hard to judge the correctness and completeness of the AWN~V1 synset  \{\AR{تحريك، تسيير، دفع، دسر}\} that corresponds to the English WordNet synset \{\textit{actuation, propulsion: the act of propelling; actuation of this app needs a password}\}.  
%In this paper, we introduce a methodology that
%enhances the content quality of WordNet. This
%method ensures synset completeness by adding
%missing lemmas, a synset gloss, and lemma examples. It also ensures synset correctness by removing irrelevant (noisy) lemmas. This method is
%verified through a large-scale case study in which
%we address 5,554 synsets covering various word
%types (nouns, verbs, adjectives, and adverbs) from
%AWN V1

In this paper, we introduce AWN~V3, a significantly extended and quality-enhanced version of AWN~V1. The novel contents of this new Arabic wordnet are: (a)~the addition of glosses and examples to all synsets; (b)~the improvement of the correctness and the completeness of the wordnet by adding missing lemmas and removing erroneous ones; (c)~a reduced level of polysemy with respect to other wordnets through the elimination of redundant word meanings, based on our prior research; and (d)~addressing phenomena of language diversity by introducing new linguistic information, namely \emph{lexical gaps} that explicitly indicate untranslatability \cite{giunchiglia2018,bella2022a} and \emph{phrasets}, i.e.,~free combinations of words that express the meaning of a synset in case of nonexistent equivalent lemmas \cite{bentivogli2000}. Such explicit representations of untranslatability distinguishes them from resource incompleteness (i.e.,~words merely missing from the resource) and give indications to both human and machine translators about particularly difficult cases of translation. Also, we tackle the polysemy problem of the source synsets by not inheriting specialization polysemy \cite{freihat2013} and compound noun polysemy \cite{freihat2015} problem in the target synsets. 

Accordingly, the paper presents the following contributions: (1)~the extension of the existing Arabic wordnet model by devices for tackling untranslatability: lexical gaps and phrasets; (2)~a development methodology for lexical databases, inscribed within the expand model, that ensures a high-quality and diversity-aware output; (3)~ AWN~V3, the new and freely available Arabic wordnet resource as described above.

The rest of the paper is organized as follows. In Section \ref{secSOA}, we introduce the state of the art of Arabic wordnets. Sections~\ref{secLangDiversity} and \ref{secPolysemy} present our contributions in addressing language diversity and excessive polysemy, respectively. In Section \ref{secMethod}, we describe our synset localization method. Section \ref{secResultigResource} presents AWN~V3, the high-quality Arabic lexical resource resulting from our work. Finally, we provide conclusions and discuss future work in Section \ref{secConclusion}.

\section{State of the Art}
\label{secSOA}

The first effort of building an Arabic wordnet was undertaken by \citet{diab2004}. She introduced an automated approach known as SALAAM (Sense Assignment Leveraging Annotations And Multilinguality) to translate synsets from PWN into standard Arabic. This translation process relied on PWN 1.7 and an English-Arabic corpus as knowledge sources. Notably, her primary focus was on translating lemmas without glosses and example sentences. This approach was evaluated using a dataset comprising 447~synsets.

AWN V1 represents the inaugural Arabic WordNet developed by \citet{elkateb2006}. The development approach closely mirrors the methodology employed in creating EuroWordnet \citeplanguageresource{vossen1999}, which consists of two phases. The first phase involves constructing a foundational core wordnet centered around \emph{base concepts} \cite{vossen1998}, while the second phase focuses on expanding the core wordnet's coverage by incorporating additional criteria. This version of AWN is aligned with PWN in terms of structure and content covering WordNet domains defined by \citet{magnini2000}. This wordnet also integrates the Suggested Upper Merged Ontology (SUMO) to provide a formal semantic framework \cite{elkateb2006}.

\begin{figure}[!ht]
\begin{center}
\includegraphics[scale=0.78]{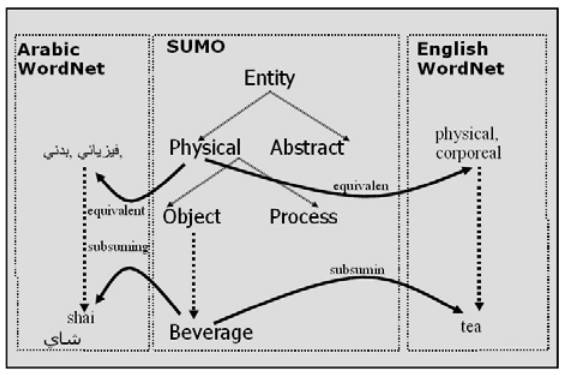} 
\caption{SUMO mapping to WordNets \cite{elkateb2006}}
\label{fig1}
\end{center}
\end{figure}

In the case of the core Arabic WordNet, the process involves encoding the Common Base Concepts (CBCs) found in the EuroWordnet and BalkaNet  \citeplanguageresource{tufis2004} as synsets. This is achieved through a manual translation effort, wherein all English synsets having an equivalence relation in the SUMO ontology are translated into their corresponding Arabic synsets. Figure \ref{fig1} illustrates this process, showing an example of how Arabic Synsets are linked to overarching SUMO terms that directly correspond to the associated English synsets. Each translated synset is validated by evaluating the coverage of synset lemmas and the domain distribution of these synsets. These efforts produced 9,228 synsets in the core wordnet of AWN V1. The distribution of these synsets, categorized by Part-Of-Speech (POS), is detailed in Table ~\ref{table1}.

\begin{table}[!ht]
\begin{center}
\begin{tabularx}{\columnwidth}{|X|c|c|c|}
    \hline
    \thead{\textbf{POS/WN}} & \thead{\textbf{AWN V1} \\ \textbf{(Core WN)}} & \thead{\textbf{AWN V1} \\ \textbf{(Ext. WN)}} & \thead{\textbf{AWN V2}} \\
    \hline
    \makecell[cl]{Noun} & 6,252 & 6,558 & 7,960 \\
    \hline
    \makecell[cl]{Verb} & 2,260 & 2,507 & 2,538 \\
    \hline
    \makecell[cl]{Adjective} & 606 & 446 & 271 \\
    \hline
    \makecell[cl]{Adverb} & 106 & 107 & 500 \\
    \hline
    \makecell[cl]{Total} & 9,228 & 9,618 & 11,269 \\
    \hline
\end{tabularx}
\caption{\label{table1}The count of Arabic synsets in each AWN version based on POS }
\end{center}
\end{table}

To expand the core of AWN, \citet{elkateb2006} introduced the Suggested Translation semi-automatic method, using available bilingual (Arabic-English) resources to extract \textit{<English word, Arabic word, POS>} tuples. This method served a similar purpose in the development of Spanish WordNet \citeplanguageresource{farreres2002} and BalkaNet \citeplanguageresource{tufis2004}. Building on eight heuristic procedures, associations between Arabic words and PWN synsets were assigned scores, relying on Arabic-English bilingual resources. Lexicographers utilized these scores to create new synsets or supplement existing ones with additional lemmas. The total number of synsets in this version is 9,618.

After the first release of AWN, there were many attempts to enrich its content concerning the number of synsets, lemma, and the relations between them. \citet{alkhalifa2009} introduced an automated method to enhance the coverage of named entities (NE) within AWN V1. This method used Wikipedia and established connections to PWN 2.0. In this study, 1,147 synsets were generated, covering 1,659 named entities across 31 general categories. In these studies, \citet{boudabous2013,batita2018} proposed a hybrid linguistic approach grounded in morphological patterns. They used Wikipedia and PWN to enrich AWN with new semantic relations. The former augmented AWN by establishing relations between nominal synsets, while the latter incorporated antonym relations.

As part of the ongoing efforts to enrich AWN, \citet{abouenour2013} introduced a semi-automatic method to increase the coverage of AWN V1. Their objective was to enhance named entities (NEs), verbs, and noun synsets. For the enrichment of NE synsets, the authors present a three-step methodology, which translates YAGO (Yet Another Great Ontology) entities \cite{suchanek2008} into Arabic instances and extracts Arabic synsets. Regarding verb synsets, the authors adopted a two-step approach inspired by \citet{rodriguez2008}. The first step involved suggesting new verbs by translating a set of verbs from VerbNet \citeplanguageresource{schuler2005} into standard Arabic. In the second step, Arabic verbs were interconnected with AWN synsets by establishing a graph connecting each Arabic verb with its corresponding English verbs in PWN. The authors employ a two-step method that detects hyponym/hypernym pairs from the web to enrich noun synsets. The overall result of this work is introducing a new version of AWN, known as AWN V2, including 11,269 synsets (for more details, see Table ~\ref{table1}).

Despite the previous efforts, which primarily focused on expanding the coverage of synset lemmas, AWN still falls short compared to other WordNets in terms of content quality. This assessment was highlighted by \citet{batita2018}, who emphasized in their research that ``AWN has very poor content in both quantity and quality levels.'' Our work focuses on the synset quality level, mainly on the synset correctness and completeness dimensions. AWN V1 marks a significant milestone for several reasons. Firstly, it encompasses the most common concepts and word senses found in PWN 2.0, ensuring a comprehensive representation in AWN V1. Secondly, its design and integration with PWN synsets facilitate cross-language usability. Finally, like other wordnets, AWN establishes a connection with SUMO, further enhancing its utility. Conversely, several issues related to synset quality have been identified in the majority of the synsets in this resource. These issues are also observed in AWN V2, as outlined below:
\begin{enumerate}
\item {All synsets lack gloss and/or illustrative examples.}
\item {Many synsets contain incorrect senses, lemmas (including incorrect word forms or repeated words), and incorrect relations between synsets.}
\item {Many synsets lack essential senses, lemmas, and necessary relations.}
\end{enumerate}

For instance, consider the following synset \{\AR{ ضجيج، ضوضاء}\} presented in AWN V1, corresponding to the English synset \{\textit{noise: sound of any kind, especially unintelligible or dissonant sound; he enjoyed the street noises}\}. In AWN V2, this synset was enriched to include \{\AR{ضجج، ضوض}\}, resulting in \{\AR{ضجيج، ضوضاء، ضجج، ضوض}\}. In this case, the synset incorporates two erroneous lemmas \{\AR{ضجج، ضوض} \}, which are not found in Arabic dictionaries such as \AR{المعاني} Almaany dictionary\footnote{\url{http://www.almaany.com/thesaurus.php}}. Additionally, it lacks the lemma \AR{ضجة}, which means \textit{noise}.
% This paper is dedicated to addressing the concerns regarding the quality of synsets within Arabic WordNet. Our primary focus is on improving the precision of synset elements and expanding the coverage of synset lemmas. Building on previous initiatives to enhance AWN, we have introduced a new version of Arabic WordNet. In our approach, we begin by utilizing the content from AWN V1 as our initial dataset, a process elaborated upon in the following section. Our goal is to identify and rectify inaccuracies and incompleteness issues.

In this paper, we enhance the accuracy of synset elements in AWN V1 by addressing incorrect lemmas and expanding the coverage of synsets through the addition of missing lemmas.

\section{Addressing Language Diversity}
\label{secLangDiversity}

Cultural and linguistic differences abound across the more than seven thousand languages in the world, to which we simply refer as language diversity. To give a few examples from lexical semantics, the English word cousin, meaning \textit{the child of your aunt or uncle}, does not have any equivalent term in Arabic. In contrast, the Arabic word \AR{عم}, which means \textit{the brother of your father}, does not exist in English \cite{Khalilia2023}. Another example is from colors: the Italian word \textit{marrone}, which means \textit{chestnut color}, does not have an equivalent word in Persian and Welsh \cite{mccarthy2019}, while the Breton \emph{glaz}, spanning a range of hues between blue and green, has no equivalent in English or in the majority of Indo-European languages.

Linguists refer to such cases of lexical untranslatability as \emph{lexical gaps}. A lexical gap happens when a word in one language is not lexicalized in another language \cite{lehrer1970}. In such cases, speakers can express a similar meaning through a free combination of words called \emph{phrasets} \cite{bentivogli2000}.

As in most wordnets, instances of language diversity are not explicitly indicated in the existing versions of AWN, instead mapping Arabic synsets to PWN synsets in an approximate manner. Such inaccuracies lead to the corruption of resource quality and an Anglo-Saxon meaning bias, also reducing the performance of applications relying on the resource, such as translation tools.

This paper introduces a new version of AWN that explicitly represents gaps and phrasets. For example, \textit{ \{adjectively: as an adjective; nouns are frequently used adjectively\}} is identified a gap in the resulting resource; at the same time, this phraset \AR{على شكل صفة} is used to describe this synset. In addition, in the case of lexicalizations (translated synsets), to increase the clarity of synset meaning and understandability, phrasets are used. For example, \textit{ \{unwittingly, unknowingly, inadvertently: without knowledge or intention; he unwittingly deleted the references \} } is translated \{ \AR{سهواً:  على نحو غير مقصود ودون إدراك أو معرفة ، حذفت الملف بدون قصد } \}, and the phraset \AR{بدون قصد} is used.

Lexical gaps are implemented in our resource at the synset level, while phrasets are implemented on the word level.

\section{Addressing Polysemy}
\label{secPolysemy}

Polysemy is a well-known problem in PWN. It has been addressed in many studies, such as \cite{gonzalo2004,mihalcea2001,buitelaar1998,freihat2014}. In our previous research \cite{freihat2016}, polysemy was classified into several types. These types are homonomy, metaphoric, metonymy, specialization polysemy, and compound noun polysemy. While the first three polysemy types are essential in lexical resources, the latter two are considered the main reasons behind the highly polysemous nature of WordNet that makes WordNet too fine-grained for NLP. As an example of compound noun polysemy, the word \textit{head} has more than 30~synsets (meanings) in PWN. Another example of compound noun polysemy is the word \textit{center}, which has 18~synsets. The problem becomes more clear in the Arabic ontology \cite{jarrar2021}, which has more than 500~synsets \AR{مركز} meaning \textit{center}. 
For example, the word \textit{turtledove} is polysemous because it belongs to the following two synsets:
\textit{ \{australian turtledove, turtledove: small Australian dove.\}}, \textit{\{turtledove: any of several Old World wild doves.\}}
Of course, it is possible to use the word \textit{turtledove} to refer to any kind of turtledoves when it is clear from the context which kind of turtledoves we are speaking about. At the same time, adding the word \textit{turtledove} as a synonym to all kinds of turtledoves in the lexical resource is useless and just makes the resource hard to use. 

According to our research \cite{freihat2015}, the word sense disambiguation for these two types is similar to anaphora
resolution and does not require including all these possible meanings in a lexical resource because they lead to the problem of sense enumeration which makes such resources very hard to use in NLP. 

\section{Addressing Synset Quality}
\label{secMethod}
In the following, we list the goals of our approach:
\begin{enumerate}
\item {Synset glosses: Each synset should have a gloss that clearly identifies its meaning. Without such gloss, we will not be able to understand the synset, moreover, we will not be able to differentiate between the meanings of the same lemma in different synsets, for example, the word \textit{'love'} has more than one meaning e.g, belongs to different synsets}
\item {Synset Examples: Each lemma in a synset should have at least one example to clarify its usage. Such examples also allow us to verify the synonymity between the synset lemmas. This is crucial for the synset correctness.}
\item {Language diversity and phrasets: Ideas are expressed in cultures in different ways, which leads to untranslatability in some languages (e.g., a lexical gap). Another phenomenon in Arabic (and maybe in other languages) is the usage of prepositional phrasets to express a synset meaning. For example, the meaning of this synset \{ \textit{someday: some unspecified time in the future; someday you will understand my actions}\} is identified as a lexical gap in Arabic, and the phraset \AR{يوماً ما} is used to express this meaning. We add these phrasets to the Arabic WordNet to increase the understandability of synsets. Also, such phrasets can be used in NLP applications to identify the intended synset.}
\item {Errors in the source WordNet (PWN): 
PWN suffers from the polysemy problem. According to our previous approaches, the source of the polysemy problem is due to the specialization polysemy and sense enumeration. In our work, we avoid such polysemy types in the resulting resource to enhance AWN usability in NLP applications.}
\item {Named entities: A lexical resource should include concepts only. It is not the correct place to include named entities, which may be another source of
 noise in lexical semantic resources.}
\end{enumerate}

Our approach consists of three steps:
\begin{enumerate}
\item{Task generation:} We have collected the data from AWN V1 and prepared the spreadsheet to be provided for translation.
\item{Task enhancement:} The translators translated the corresponding PWN synset glosses, then performed the following: adding missing lemmas, and examples for the synset elmmas, removing wrong lemmas from the original Arabic synsets, identifying gaps in the case of untranlatability, and adding phrasets for increasing the understandability of synsets. 
\item{Task Validation}: Validation is carried out in two phases: 1) Each contribution provided by one of the translators was validated by the other. In the second phase, a linguistic expert validates and approves the contribution.
\end{enumerate}

% In this section, we outline the methodology employed to enhance the quality of the Arabic WordNet and use it to introduce an improved version of the Arabic WordNet. This involved the collection of missing synset lemmas, glosses, and example sentences, alongside tackling irrelevant (incorrect or duplicated) lemmas within synsets, in collaboration with translators and a linguistic expert. The insights provided in this section are intended to serve as a well-established framework for elevating the content quality of AWN V1, which we envisage employing in future enhancements of AWN V2.
% We use AWN V1 (the expanded version) as our source for importing Arabic synsets, which may require quality enhancements, to serve as the input dataset for our method. Also, the data representation model of AWN V1 is adopted, which is intricately linked with the PWN hierarchy, facilitating the formalization of our collected data. Using this hierarchy, we understand the exact meaning of each synset, which is the initial phase in our work to enhance synset quality. Subsequently, based on the meaning of the synset, our approach enables us to systematically assess and address issues related to synset elements, such as the incorporation of any missing lemmas. This method consists of two steps: Firstly, a linguistic expert generates a contribution task to prepare needed material, such as input data. Secondly, the translators are asked to evaluate the input data and provide enhancements using PWN as a reference.  Accordingly, we present the steps of our methodology in detail as follows:

\subsection{Task generation}

This section describes the essential materials required for the next step of the methodology. The preparation process involves constructing a dataset containing AWN V1 synsets as well as the corresponding PWN synsets. In this context, AWN V1 and PWN browsers are utilized for data retrieval. This dataset is customized in a spreadsheet for usability and simplicity in providing contributions, in which the linguistic expert (the first author) organizes synsets into four categories (each in one sheet) based on the part of speech (POS). Each row within the spreadsheet represents a synset and includes information such as the synset ID, lemmas, gloss, and example sentences in Standard Arabic and English. Additionally, empty slots are provided for inserting missing lemmas, a gloss, examples, and comments by the data provider (translator) in Arabic. One additional slot is designated for validation purposes, along with comments from the validator. In this step, the linguistic expert excludes all (42 synsets) named entities from the spreadsheet.

\subsection{Task enhancement}

\begin{figure}[!ht]
\begin{center}
\includegraphics[scale=0.60]{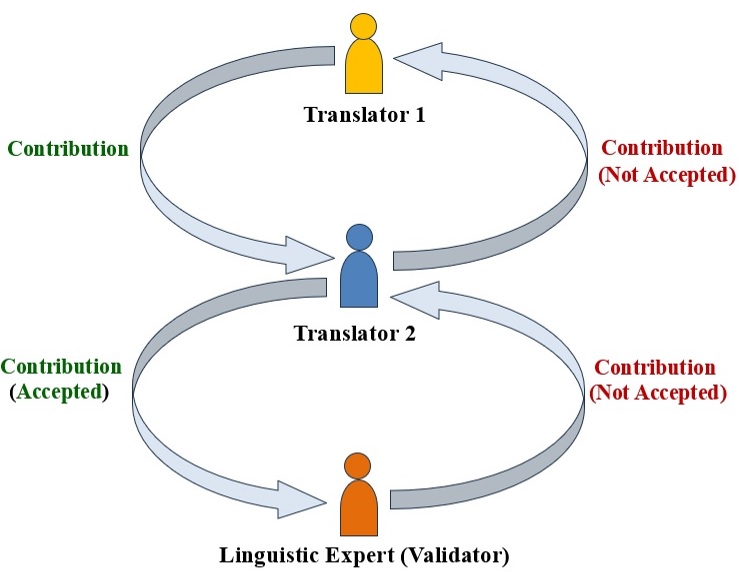} 
\caption{The workflow of the contribution collection}
\label{fig2}
\end{center}
\end{figure}

Contributions for synset enhancement, which involve the addition of missing information or correction of synset elements, are made by two translators and validated by a language expert. An overview of our contribution collection workflow is illustrated in Figure \ref{fig2}. As depicted, the workflow is structured into two cycles, with the aim of ensuring the quality of results. The first cycle operates between the two translators, where each translator's contributions are subject to verification by the other. The second cycle involves the validation of accepted contributions by a linguistic expert. 

% The process of gathering contributions starts with providing the translators with a spreadsheet created in the previous step. Then, the translators are tasked to translate each PWN synset to Arabic or identify it as a lexical gap using a bilingual (English-Arabic) linguistic resource, such as the Al-Mawrid Al-Qareeb \AR{المورد القريب} dictionary \cite{baalbaki2005}. After that, by comparing a generated Arabic synset with the corresponding Arabic synset in the spreadsheet, they tackle the Arabic synset (in the spreadsheet) by adding missing synset items and/or rectifying incorrect elements or marking the synset as a gap in Arabic and providing a free combination of words to express the synset.

The process of synset enhancement in the first cycle was carried out by two native speakers. Regarding their socio-linguistic background, both translators possess at least a bachelor’s degree in the field of translation (English-Arabic). Before the translation, translators have been trained as described in the following subsections.

\subsubsection{Synset understanding}

Central to this process is ensuring that the translator possesses a clear understanding of the synset they are tasked with translating. Misunderstandings can arise when the translator does not grasp a thorough understanding of both the synset lemmas and the gloss in English. The translators are asked to understand each PWN synset in the spreadsheet using the following notable instructions:
\begin{itemize}
     \item{Use external resources such as dictionaries and Wikipedia to understand the meaning of the synset.}
     \item {They are given the authority to skip the synset or leave a comment when they do not understand the meaning of the synset.}
     
\end{itemize}

\subsubsection{Lexical gap identification and synset lexicalisation}
A lexical gap happens when either the meaning of the concept in a source language is not known in the culture of the target language or the concept can be lexicalized only through word-free combination \cite{giunchiglia2018}. This means that there is no lexical unit (single word or restricted collocation) that corresponds to any of the source language lemmas. In this step, for each English synset in the spreadsheet, the translator decides whether it has an equivalent meaning in Arabic (lexicalisation exists) or is a lexical gap based on the understanding of the English synset and using a bilingual dictionary. If an English synset$_{i}$ is a gap, the translator performs \textbf{step A}; otherwise, she/he performs \textbf{steps B} and \textbf{C}.
\\

\textit{\textbf{Step A: Lexical gap processing}},
in this step, the translator is asked to mark the English synset$_{i}$ as a lexical gap in the spreadsheet and provide a phraset in Arabic. For example, the synset \{\textit{expressively: with expression, in an expressive manner; she gave the order to the waiter, using her hands very expressively}\} is identified as a lexical gap in Arabic, and \AR{بشكل معبر} meaning (an expressive way) is provided as a phraset to this synset.
\\

\textit{\textbf{Step B: Synset translation}}, after the translator confirms the existence of the meaning of the English synset$_{i}$ in Arabic, she/he translates this synset to Arabic. This translation includes the following steps.

\begin{enumerate}
\item { \textbf{Translating synset gloss:} The translation is across language and cross-cultural communication. A translation should give a complete transcript of the synset; meanwhile, the style and manner of writing should be at least the same quality as the gloss of English. Above all, faithfulness, expressiveness, and closeness are the important three elements of translation. The gloss should explicitly express the semantics and the common attributes of a synset. }
% The semantic relation between the synset and other synsets, which is expressed in the gloss, affects the gloss writing style. In the following example, we show how the gloss expresses the hypernmy/hyponomy (is a) semantic relation.\\\textbf{Hypernymy/hyponomy} is an inheritance relation, so the child concept is a sub-concept of its parent concept. The gloss of the child concept expresses this relation in terms of genus and differentia. The genus expresses the shared properties between the parent and the child concepts, and the differentia expresses the specific properties of the child concept. For instance, the following two synsets (a) and (b) share the same parent school, so the genus of both synsets is school, and the differentia is the other parts besides school. So, in this example, the relation is encoded through this pattern: \textit{Gloss = genus + differentia}.\\(a) \{driving school: a school where people are taught to drive automobiles\} \\(b) \{flying school: a school for teaching students to fly airplanes\}}
\item {\textbf{Translating synset lemmas:} Translators should keep two key considerations in mind while translating synset lemmas. Firstly, this translation process does not entail a direct one-to-one correspondence between English and Arabic terms. Secondly, it is important to note that the set of lemmas within the English synset may not be exhaustive, meaning it might not contain all the synonyms associated with the synset. To translate the synset lemmas, we go through the following phases:}
% \end{enumerate}
\begin{itemize}
\item {\textbf{English Lemmas translation:} Translate the English synset lemmas into Arabic. The result of this step is a set of lemmas of the length $n$, where $n$ is the number of lemmas in the English synset.}
\item {\textbf{Arabic synonyms collection:} For each translated lemma, the translators collect the lemma synonyms in Arabic. The result of this phase is m synonym sets in Arabic, $m\leq n$ (since some Arabic lemmas may have empty synonym sets).}
\item {\textbf{Arabic Synonyms validation:} Based on the synset gloss, for each of the m synonym sets in Arabic, the translators exclude all synonyms that do not belong to the synset. Use the provided examples in the English synset and other examples to include/exclude the synonyms in this phase.} 
% For example, the lemma \AR{مبنى المدرسة} meaning \textit{school building} does not belong to this synset \{ \AR{ فترة التدريس في المدرسة أو وقت الحصص التدريسية :المدرسة ، وقت المدرسة ، يوم مدرسي} \} which the translated of this synset \{ \textit{school, schooltime, school day: the period of instruction in a school; the time period when schools is in session} \}.}
\item {\textbf{Arabic lemmas collection:} The translators collect the Arabic lemmas, resulting in the translation process in phase (1) and the synonyms produced from phase (2) and put them as the Arabic synset lemmas. In the case of polysemy, we solve the specialization polysemy and compound noun polysemy. For example, \AR{جِسْم}  is excluded from this synset \{\AR{ جِسْم فِيزْيائِي، جِسْم}\} which corresponds \{\textit{object, physical object: a tangible and visible entity, an entity that can cast a shadow;	pens, books and bags are school objects}\}

}

\item {\textbf{Arabic lemmas ordering:} The translators order the Arabic collected lemmas in phase (4), wherein the first lemma is the Arabic synset preferred term and so on (in descending order of importance). Based on the examples provided in phase (3) (and other examples if needed), the translator gives preferences for the lemmas based on these examples. }
 % For instance, the synset \{ \AR{ احصل على المتعة من: احب، استمتع } \} corresponding to \{\textit{love, enjoy:  get pleasure from}\} and using this synset example \AR{ أنا احب الطبخ} meaning \textit{I love cooking} the translator gives the correct preference of the lemmas.} }
\end{itemize}

\item { \textbf{Translating synset examples:} Examples within a synset contribute to a clearer comprehension of how to utilize the synset lemmas, consequently enhancing the overall understanding of the fully lexicalized synset. We employ the same examples crafted during the lemma translation phase as synset examples. This approach signifies that we do not solely translate the examples found in the English synset. Ideally, we provide an example sentence in Arabic for each synonym within the synset, even if the English synset does not contain examples at all. The provided examples are incorporated into the Aarbic synsets, aligning them with the order of their respective synonyms}

\end{enumerate}

\textit{\textbf{Step C: Comparing the produced (translated) synset in Step (B) with the corresponding synset from AWN V1}}, At this stage, a translator compares the translated synset generated in Step B and its corresponding Arabic synset, as imported from AWN V1. This Arabic synset is designated to correspond with the English synset$_{i}$ in the spreadsheet. Based on the gloss and examples provided in the generated synset, the translators undertake the following actions: (1) Copy lemmas from the translated synset to the AWN V1 synset if they are missing from the AWN V1 synset. (2) Exclude the lemmas from the AWN V1 synset, which are not covered by the synset gloss and examples. (3) Copy the gloss and the examples from the translated synset to the AWN v1 synset if they are missing in the latter.
\\

\subsection{Task validation}
\label{secValid}
The validation process consists of two phases. In the first phase, the two translators validate the resulting synsets (stored in a spreadsheet containing English and produced Arabic synsets) in an alternating manner, checking each synset (and gap) one by one. During the validation, each of them considers the following:
\begin{enumerate}
   \item{Gap validation}: A translator validates synsets marked as lexical gaps in Arabic, either as confirmed gaps or as non-gaps due to an existing lexicalization in Arabic, which he/she should provide a gloss and lemmas of that synset.
   
    \item{Gloss validation:} The Arabic gloss expresses the intended meaning of the English synset. Also, the Arabic gloss is easy to understand and does not contain typos or grammatical errors. 
    \item{Lemmas validation:} Synset lemmas should be correct (e.g., not include wrong lemmas) and complete (e.g., there are no missing lemmas). In addition, the validator can use the examples to check synonymity between lemmas.
    \item{Examples validation:} Each lemma has at least one example. The examples are natural and express the intended usage.   
\end{enumerate}

In case of disagreement, the affected synsets are sent back to the translators with the validator's comment. The accepted synsets are sent to the expert validation.

In the second phase, An Arabic linguistic expert performs this validation on a spreadsheet containing the resulting synsets (and gaps) only, without including the English synsets, which both translators accepted in the previous step. His task is to approve the final resulting synsets. The same criteria used in the previous validation phase for validating gaps, glosses, lemmas, and examples are adopted in this step.

% Differences: performed an Arabic language expert. the validation is performed on the Arabic synsets only (gloss in Arabic).
% 1. Does the gloss correspond to a concept?
% 2. Also, the Arabic gloss is easy to understand, which does not contain typos or grammatical errors.
% 3. in the case of a gap or not. if not he should provide the gloss and lemmas of that synset.

% The linguistic expert (the first author) confirms the accuracy and the comprehensiveness of the lemmas for the generated synset and ensures that the order of lemmas correctly represents the synset's intended preference}

\section{Evaluation and the Resulting Resource}
\label{secResultigResource}

This section demonstrates the use of the methodology described in Section \ref{secMethod} on evaluating and improving the content quality of AWN V1 depending on PWN as a reference to our work. As mentioned above, AWN V1 includes 9,618 synsets written in Modern Standard Arabic (MSA), which refers to the standard form of the language used in academic writing, formal communication, classical poetry, and religious sermons.

In this study, contributions are provided by two translators (each is an Arabic native speaker). They were born and educated within the Arabic-speaking community, having completed at least their high school education within this community.

Four experiments (one for each POS) are performed to evaluate the extended version of AWN V1 synsets and tackle synset quality issues using our method. In each experiment, a spreadsheet includes Arabic synsets imported from the AWN V1 and their corresponding English synsets. Each spreadsheet contains data for a specific POS and serves as an input dataset to the contribution (synset quality enhancement) collection step. The experiments are conducted on 6,516~nouns, 2,507~verbs, 446~adjectives, and 107~adverbs (see Table~\ref{experiment} for more details).

\begin{table}[!ht]
\small
\setlength{\tabcolsep}{4pt}
\begin{center}
\begin{adjustbox}{width=.5\textwidth,center}
\begin{tabular}{|l|c|c|c|c|c|}
    \hline
    & \thead{\textbf{Noun}} & \thead{\textbf{Verb}} & \thead{\textbf{Adjective}} & \thead{\textbf{Adverb}} & \thead{\textbf{Total}} \\
    \hline
    \makecell[cl]{Synsets} & 6,516 & 2,507 & 446 & 107 & \textbf{9,576} \\
    \hline
    \makecell[cl]{Words} & 13,659 & 5,878 & 761 & 262 & \textbf{20,560} \\
    \hline
            
\end{tabular}
\end{adjustbox}
\caption{\label{experiment}The count of synsets and words (imported from the extended AWN V1- without named entities) in the input dataset based on POS}
\end{center}
\end{table}

In the contribution collection, for each Arabic synset in a row in the spreadsheet, a translator is tasked to translate the corresponding PWN synset to Arabic or identify it as a lexical gap using a bilingual (English-Arabic) linguistic resource, such as the Al-Mawrid Al-Qareeb \AR{المورد القريب} dictionary \cite{baalbaki2005}. After that, if a lexicalization exists in Arabic, the translator tackles the latter by comparing a generated translated Arabic synset with the AWN V1 synset in the same row, which follows by adding missing synset lemmas, gloss, and example sentences; and/or rectifying incorrect elements. Also, if the English synset is a gap in Arabic, he/she marks it as a lexical gap and provides a phraset to express the synset (Note that phraset is also used for some translated synsets to increase the understandability). To our knowledge, our resulting resource (AWN~V3) is the first Arabic Wordnet that identifies gaps and provides phrasets.

The overall effort to collect contributions resulted in updating 5,554~synsets from AWN~V1. We added 2,726~new lemmas, 9,322~new glosses, and 12,204~new example sentences. We also identified 236~lexical gaps and inserted 701~phrasets. Furthermore, we deleted 8751~incorrect lemmas. More details regarding the counts of these contributions are presented in Table ~\ref{table3}. See the dataset uploaded to GitHub\footnote{\url{https://github.com/HadiPTUK/AWN3.0}}. For each POS, two spreadsheets were uploaded to GitHub; the first file includes the final resulting Arabic synsets, and the second contains the added and deleted synset components.
%\footnote{\url{https://anonymous.4open.science/r/AWN3}}

\begin{table}[!ht]
\small
\setlength{\tabcolsep}{4pt}
\begin{center}
\begin{adjustbox}{width=.5\textwidth,center}
\begin{tabular}{|l|c|c|c|c|c|}
    \hline
    & \thead{\textbf{Noun}} & \thead{\textbf{Verb}} & \thead{\textbf{Adj}} & \thead{\textbf{Adv}} & \thead{\textbf{Total}} \\
    \hline
    \makecell[cl]{Updated synsets} & 3,938 & 1,364 & 181 & 71 & \textbf{5,554} \\
    \hline
    \makecell[cl]{New lemmas} & 2,581 & 64 & 72 & 9 & \textbf{2,726} \\
    \hline
    \makecell[cl]{Deleted lemmas} & 6,050 & 2,387 & 223 & 91 & \textbf{8,751} \\
    \hline
    \makecell[cl]{New glosses} & 6,511 & 2,258 & 446 & 107 & \textbf{9,322} \\
    \hline
    \makecell[cl]{New examples} & 7,597 & 3,620 & 782 & 205 & \textbf{12,204} \\
    \hline
    \makecell[cl]{Gaps} & 28 & 187 & 0 & 21 & \textbf{236} \\
    \hline
    \makecell[cl]{Phrasets} & 364 & 275 & 0 & 62 & \textbf{701} \\
    \hline
    
\end{tabular}
\end{adjustbox}
\caption{\label{table3}Statistics of the data addition and deletion into/from AWN}
\end{center}
\end{table}

Validation was carried out by an Arabic linguistic expert who has a Ph.D. in the Arabic language and is a university instructor at the linguistics department. As introduced above, the expert follows the criteria described in Section \ref{secValid} to verify produced synsets. Results can be seen in Table~\ref{validation}, where by correctness we understand the number of contributions validated as correct divided by the total number of contributions. These contributions can be newly added or deleted lemmas, collected glosses and example sentences, identified lexical gaps, and inserted phrasets.    
For example, in the case of an added lemma, the validator either confirms the addition or rejects it by leaving a comment. For instance, \{\AR{مِقْيَاس}\} meaning \textit{a measuring tool} is deemed an incorrect added word to the synset \{\AR{ مِقْدَار، قَدْر، كَمّ، كَمِّيَّة}\} which corresponds to \textit{\{measure, quantity, amount: how much there is of something that you can quantify; he has a big amount of money\}}. In the case of identified gaps, the validator either as confirmed gaps or as non-gaps due to an existing lexicalization in Arabic, which the validator needs to indicate. For instance, the following English synset \textit{ \{try, try on: put on a garment in order to see whether it fits and looks nice; Try on this sweater to see how it looks\} } is considered a gap. The validator rejected it and provided this word \AR{قَاس} with the same meaning.

\begin{table}[!ht]
\begin{center}
\begin{tabularx}{\columnwidth}{|X|Y|}
    \hline
    \thead{\textbf{Contribution}} & \thead{\textbf{Correctness}} \\
    \hline
    \makecell[cl]{New lemmas} & 97.34\% \\
    \hline
    \makecell[cl]{Deleted lemmas} & 98.89\% \\
    \hline
    \makecell[cl]{New glosses} & 98.76\% \\
    \hline
    \makecell[cl]{New examples} & 99.13\% \\
    \hline
    \makecell[cl]{Gaps} & 96.82\% \\
    \hline
    \makecell[cl]{Phrasets} & 97.54\% \\
    \hline
    \makecell[cl]{Total} & 98.08\%\\
    \hline
\end{tabularx}
\caption{\label{validation}Validator evaluation of translator contributions}
\end{center}
\end{table}

Upon discussion between the validator (linguistic expert) and the translators, the mistakes made by the latter can be explained by misunderstandings of the meanings of certain concepts provided in English. The validator made sure to exclude or fix the mistakes, bringing the correctness of the final dataset closer to 100\%.

\section{Conclusion and Future Work} 
\label{secConclusion}

In this paper, we evaluate and address the quality---correctness and completeness---of synsets from AWN V1 across four parts of speech (nouns, verbs, adjectives, and adverbs). The resulting total of 9,576~synsets are introduced as AWN~V3---an enhanced version of AWN with corrected and extended lemmas, as well as added glosses and example sentences. In order to represent English words not directly translatable to Arabic, we introduce \emph{phrasets} to provide approximate phrase-level translations and \emph{lexical gaps} to indicate untranslatability. As part of our future work, we will apply the methodology described in order to increase the coverage of Arabic synsets, based on AWN~V2 as well as the remaining synsets in PWN.

%\section{Acknowledgements} We thank the University of Trento and Palestine Technical University—Kadoori for their support.

\nocite{*}
\section{Bibliographical References}
\label{sec:reference}
\bibliographystyle{lrec-coling2024-natbib}
\bibliography{lrec-coling2024-example}

\begin{thebibliography}{9}
\expandafter\ifx\csname natexlab\endcsname\relax\def\natexlab#1{#1}\fi

\bibitem[{Bhattacharyya(2010)}]{bhattacharyya2010}
Pushpak Bhattacharyya. 2010.
\newblock \href {http://www.lrec-conf.org/proceedings/lrec2010/pdf/939_Paper.pdf} {{I}ndo{W}ord{N}et}.
\newblock In \emph{Proceedings of the Seventh International Conference on Language Resources and Evaluation ({LREC}'10)}, Valletta, Malta. European Language Resources Association (ELRA).

\bibitem[{Elkateb et~al.(2006)Elkateb, Black, Rodr{\'\i}guez, Alkhalifa, Vossen, Pease, and Fellbaum}]{elkateb2006}
Sabri Elkateb, William Black, Horacio Rodr{\'\i}guez, Musa Alkhalifa, Piek Vossen, Adam Pease, and Christiane Fellbaum. 2006.
\newblock Building a {W}ord{N}et for {A}rabic.
\newblock In \emph{Proceedings of the Fifth International Conference on Language Resources and Evaluation ({LREC}{'}06)}, pages 29--34. European Language Resources Association.

\bibitem[{Farreres et~al.(2002)Farreres, Rodr{\'\i}guez, and Gibert}]{farreres2002}
Javier Farreres, Horacio Rodr{\'\i}guez, and Karina Gibert. 2002.
\newblock Semiautomatic creation of taxonomies.
\newblock In \emph{COLING-02: SEMANET: Building and Using Semantic Networks}.

\bibitem[{Miller(1995)}]{miller1995}
George~A Miller. 1995.
\newblock Wordnet: a lexical database for {E}nglish.
\newblock \emph{Communications of the ACM}, 38(11):39--41.

\bibitem[{Piasecki et~al.(2009)Piasecki, Broda, and Szpakowicz}]{piasecki2009}
Maciej Piasecki, Bernd Broda, and Stanislaw Szpakowicz. 2009.
\newblock \emph{A {W}ordnet from the ground up}.
\newblock Oficyna Wydawnicza Politechniki Wroc{\l}awskiej Wroc{\l}aw.

\bibitem[{Regragui et~al.(2016)Regragui, Abouenour, Krieche, Bouzoubaa, and Rosso}]{regragui2016}
Yasser Regragui, Lahsen Abouenour, Fettoum Krieche, Karim Bouzoubaa, and Paolo Rosso. 2016.
\newblock Arabic {W}ordnet: New content and new applications.
\newblock In \emph{Proceedings of the 8th Global WordNet Conference (GWC)}, pages 333--341.

\bibitem[{Schuler(2005)}]{schuler2005}
Karin~Kipper Schuler. 2005.
\newblock \emph{VerbNet: A broad-coverage, comprehensive verb lexicon}.
\newblock University of Pennsylvania.

\bibitem[{Tufis et~al.(2004)Tufis, Cristea, and Stamou}]{tufis2004}
Dan Tufis, Dan Cristea, and Sofia Stamou. 2004.
\newblock Balkanet: Aims, methods, results and perspectives. a general overview.
\newblock \emph{Romanian Journal of Information science and technology}, 7(1-2):9--43.

\bibitem[{Vossen(1999)}]{vossen1999}
PJTM Vossen. 1999.
\newblock Eurowordnet.

\end{thebibliography}


\begin{thebibliography}{31}
\expandafter\ifx\csname natexlab\endcsname\relax\def\natexlab#1{#1}\fi

\bibitem[{Abouenour et~al.(2013)Abouenour, Bouzoubaa, and Rosso}]{abouenour2013}
Lahsen Abouenour, Karim Bouzoubaa, and Paolo Rosso. 2013.
\newblock On the evaluation and improvement of {A}rabic {W}ordnet coverage and usability.
\newblock \emph{Language resources and evaluation}, 47:891--917.

\bibitem[{Alkhalifa and Rodr{\'\i}guez(2009)}]{alkhalifa2009}
Musa Alkhalifa and Horacio Rodr{\'\i}guez. 2009.
\newblock Automatically extending {NE} coverage of {A}rabic {W}ordnet using {W}ikipedia.
\newblock In \emph{Proc. Of the 3rd International Conference on Arabic Language Processing CITALA2009, Rabat, Morocco}, pages 23--30.

\bibitem[{Baalbaki(2005)}]{baalbaki2005}
Rohi Baalbaki. 2005.
\newblock \emph{Al-mawrid Al-qareeb {A}rabic-{E}nglish Dictionary}.
\newblock Dar El Ilm Lilmalayin, Lebanon.

\bibitem[{Batita and Zrigui(2018)}]{batita2018}
Mohamed~Ali Batita and Mounir Zrigui. 2018.
\newblock The enrichment of {A}rabic {W}ordnet antonym relations.
\newblock In \emph{Computational Linguistics and Intelligent Text Processing: 18th International Conference, CICLing 2017, Budapest, Hungary, April 17--23, 2017, Revised Selected Papers, Part I 18}, pages 342--353. Springer.

\bibitem[{Beckwith et~al.(2021)Beckwith, Fellbaum, Gross, and Miller}]{beckwith2021}
Richard Beckwith, Christiane Fellbaum, Derek Gross, and George~A Miller. 2021.
\newblock Wordnet: A lexical database organized on psycholinguistic principles.
\newblock In \emph{Lexical Acquisition}, pages 211--232. Psychology Press.

\bibitem[{Bella et~al.(2022)Bella, Batsuren, Khishigsuren, and Giunchiglia}]{bella2022a}
G{\'a}bor Bella, Khuyagbaatar Batsuren, Temuulen Khishigsuren, and Fausto Giunchiglia. 2022.
\newblock Linguistic diversity and bias in online dictionaries.
\newblock \emph{University of Bayreuth African Studies Online}, page 173.

\bibitem[{Ben{\'\i}tez et~al.(1998)Ben{\'\i}tez, Cervell, Escudero, L{\'o}pez, Rigau, and Taul{\'e}}]{benitez1998}
Laura Ben{\'\i}tez, Sergi Cervell, Gerard Escudero, M{\`o}nica L{\'o}pez, German Rigau, and Mariona Taul{\'e}. 1998.
\newblock Methods and tools for building the {C}atalan {W}ordnet.
\newblock \emph{arXiv preprint cmp-lg/9806009}.

\bibitem[{Bentivogli and Pianta(2000)}]{bentivogli2000}
Luisa Bentivogli and Emanuele Pianta. 2000.
\newblock Looking for lexical gaps.
\newblock In \emph{Proceedings of the ninth EURALEX International Congress}, pages 8--12. Stuttgart: Universit{\"a}t Stuttgart.

\bibitem[{Boudabous et~al.(2013)Boudabous, Kammoun, Khedher, Belguith, and Sadat}]{boudabous2013}
Mohamed~Mahdi Boudabous, Nouha~Cha{\^a}ben Kammoun, Nacef Khedher, Lamia~Hadrich Belguith, and Fatiha Sadat. 2013.
\newblock Arabic {W}ordnet semantic relations enrichment through morpho-lexical patterns.
\newblock In \emph{2013 1st International Conference on Communications, Signal Processing, and their Applications (ICCSPA)}, pages 1--6. IEEE.

\bibitem[{Buitelaar(1998)}]{buitelaar1998}
Peter~Paul Buitelaar. 1998.
\newblock \emph{CoreLex: systematic polysemy and underspecification}.
\newblock Brandeis University.

\bibitem[{Diab(2004)}]{diab2004}
Mona Diab. 2004.
\newblock The feasibility of bootstrapping an {A}rabic {W}ordnet leveraging parallel corpora and an {E}nglish {W}ordnet.
\newblock In \emph{Proceedings of the Arabic Language Technologies and Resources, NEMLAR, Cairo}.

\bibitem[{Freihat(2014)}]{freihat2014}
Abed~Alhakim Freihat. 2014.
\newblock \emph{An organizational approach to the polysemy problem in {W}ordnet}.
\newblock Ph.D. thesis, University of Trento.

\bibitem[{Freihat et~al.(2013)Freihat, Giunchiglia, and Dutta}]{freihat2013}
Abed~Alhakim Freihat, Fausto Giunchiglia, and Biswanath Dutta. 2013.
\newblock Solving specialization polysemy in wordnet.
\newblock \emph{Int. J. Comput. Linguistics Appl.}, 4(1):29--52.

\bibitem[{Freihat et~al.(2016)Freihat, Giunchiglia, and Dutta}]{freihat2016}
Abed~Alhakim Freihat, Fausto Giunchiglia, and Biswanath Dutta. 2016.
\newblock A taxonomic classification of {W}ordnet polysemy types.
\newblock In \emph{Proceedings of the 8th Global WordNet Conference (GWC)}, pages 106--114.

\bibitem[{Freihat et~al.(2015)Freihat, Dutta, and Giunchiglia}]{freihat2015}
Abed~Alhkaim Freihat, Biswanath Dutta, and Fausto Giunchiglia. 2015.
\newblock Compound noun polysemy and sense enumeration in {W}ordnet.
\newblock In \emph{Proceedings of the 7th International Conference on Information, Process, and Knowledge Management (eKNOW)}, pages 166--171.

\bibitem[{Giunchiglia et~al.(2018)Giunchiglia, Batsuren, and Alhakim~Freihat}]{giunchiglia2018}
Fausto Giunchiglia, Khuyagbaatar Batsuren, and Abed Alhakim~Freihat. 2018.
\newblock One world-seven thousand languages (best paper award, third place).
\newblock In \emph{International Conference on Computational Linguistics and Intelligent Text Processing}, pages 220--235. Springer.

\bibitem[{Gonzalo(2004)}]{gonzalo2004}
Julio Gonzalo. 2004.
\newblock Sense proximity versus sense relations.
\newblock \emph{GWC 2004}, page~5.

\bibitem[{Jarrar(2021)}]{jarrar2021}
Mustafa Jarrar. 2021.
\newblock The {A}rabic ontology--an {A}rabic {W}ordnet with ontologically clean content.
\newblock \emph{Applied ontology}, 16(1):1--26.

\bibitem[{Khalilia et~al.(2023)Khalilia, Bella, Freihat, Darma, and Giunchiglia}]{Khalilia2023}
Hadi Khalilia, Gábor Bella, Abed~Alhakim Freihat, Shandy Darma, and Fausto Giunchiglia. 2023.
\newblock \href {https://doi.org/10.3389/fpsyg.2023.1229697} {Lexical diversity in kinship across languages and dialects}.
\newblock \emph{Frontiers in Psychology}, 14.

\bibitem[{Khalilia et~al.(2021{\natexlab{a}})Khalilia, Freihat, and Giunchiglia}]{khalilia2021a}
Hadi Khalilia, Abed~Alhakim Freihat, and Fausto Giunchiglia. 2021{\natexlab{a}}.
\newblock The quality of lexical semantic resources: A survey.
\newblock In \emph{Proceedings of the 4th International Conference on Natural Language and Speech Processing (ICNLSP 2021)}, pages 117--129.

\bibitem[{Khalilia et~al.(2021{\natexlab{b}})Khalilia, Freihat, Giunchiglia et~al.}]{khalilia2021b}
Hadi Khalilia, Abed~Alhakim Freihat, Fausto Giunchiglia, et~al. 2021{\natexlab{b}}.
\newblock The dimensions of lexical semantic resource quality.
\newblock In \emph{Proceedings of the Second International Workshop on NLP Solutions for Under Resourced Languages (NSURL 2021) co-located with ICNLSP 2021}, pages 15--21. ACL Anthology.

\bibitem[{Lehrer(1970)}]{lehrer1970}
Adrienne Lehrer. 1970.
\newblock Notes on lexical gaps.
\newblock \emph{Journal of linguistics}, 6(2):257--261.

\bibitem[{Magnini and Cavaglia(2000)}]{magnini2000}
Bernardo Magnini and Gabriela Cavaglia. 2000.
\newblock Integrating subject field codes into {W}ordnet.
\newblock In \emph{LREC}, volume 1413.

\bibitem[{McCarthy et~al.(2019)McCarthy, Wu, Mueller, Watson, and Yarowsky}]{mccarthy2019}
Arya~D McCarthy, Winston Wu, Aaron Mueller, Bill Watson, and David Yarowsky. 2019.
\newblock Modeling color terminology across thousands of languages.
\newblock \emph{arXiv preprint arXiv:1910.01531}.

\bibitem[{Mihalcea and Moldovan(2001)}]{mihalcea2001}
Rada Mihalcea and Dan~I Moldovan. 2001.
\newblock Ez. {W}ordnet: Principles for automatic generation of a coarse grained wordnet.
\newblock In \emph{FLAIRS conference}, pages 454--458.

\bibitem[{Navigli(2009)}]{navigli2009}
Roberto Navigli. 2009.
\newblock Word sense disambiguation: A survey.
\newblock \emph{ACM computing surveys (CSUR)}, 41(2):1--69.

\bibitem[{Nie(2022)}]{nie2022}
Jian-Yun Nie. 2022.
\newblock \emph{Cross-language information retrieval}.
\newblock Springer Nature.

\bibitem[{Poibeau(2017)}]{poibeau2017}
Thierry Poibeau. 2017.
\newblock \emph{Machine translation}.
\newblock MIT Press.

\bibitem[{Rodr{\'\i}guez et~al.(2008)Rodr{\'\i}guez, Farwell, Ferreres, Bertran, Alkhalifa, and Mart{\'\i}}]{rodriguez2008}
Horacio Rodr{\'\i}guez, David Farwell, Javi Ferreres, Manuel Bertran, Musa Alkhalifa, and Maria~Ant{\`o}nia Mart{\'\i}. 2008.
\newblock Arabic wordnet: Semi-automatic extensions using bayesian inference.
\newblock In \emph{LREC}.

\bibitem[{Suchanek et~al.(2008)Suchanek, Kasneci, and Weikum}]{suchanek2008}
Fabian~M Suchanek, Gjergji Kasneci, and Gerhard Weikum. 2008.
\newblock Yago: A large ontology from {W}ikipedia and {W}ordnet.
\newblock \emph{Journal of Web Semantics}, 6(3):203--217.

\bibitem[{Vossen(1998)}]{vossen1998}
Piek Vossen. 1998.
\newblock A multilingual database with lexical semantic networks.
\newblock \emph{Dordrecht: Kluwer Academic Publishers. doi}, 10:978--94.

\end{thebibliography}

\section{Language Resource References}
\label{lr:ref}
\bibliographystylelanguageresource{lrec-coling2024-natbib}
\bibliographylanguageresource{languageresource}

\end{document}